\documentclass[twoside]{article}

\usepackage[accepted]{aistats2024}
\usepackage{microtype}
\usepackage{graphicx}
\usepackage{subfigure}
\usepackage{booktabs}
\usepackage{hyperref}
\usepackage{amsmath,amssymb,amsfonts,mathtools,amsthm,bm}
\usepackage{algorithm}
\usepackage{algpseudocode}
\usepackage{xcolor}
\usepackage{multirow}
\usepackage{enumitem}
\setlist[itemize]{align=parleft,left=0pt..1em}
\newtheorem{theorem}{Theorem}
\usepackage[round]{natbib}
\renewcommand{\bibname}{References}
\renewcommand{\bibsection}{\subsubsection*{\bibname}}

\begin{document}

%

%

\twocolumn[

\aistatstitle{Sample Efficient Learning of Factored Embeddings of Tensor Fields}

\aistatsauthor{ Taemin Heo \And Chandrajit Bajaj }

\aistatsaddress{ Department of Computer Science and Oden Institute of Computational Engineering and Sciences \\ University of Texas at Austin, Texas, USA }

]

\begin{abstract}
Data tensors of orders 2 and greater are now routinely being generated. These data collections are increasingly huge and growing. Many scientific and medical data tensors are tensor fields (e.g., images, videos, geographic data) in which the spatial neighborhood contains important information.  Directly accessing such large data tensor collections for information has become increasingly prohibitive. We learn approximate full-rank and compact tensor sketches with decompositive representations  providing  compact space, time and spectral  embeddings of tensor fields. All information querying and post-processing on the original tensor field can now be achieved more efficiently and with customizable accuracy as they are performed on these compact factored sketches in latent generative space. We produce  optimal rank-$\bm{r}$ sketchy Tucker decomposition of arbitrary order data tensors by building compact factor matrices from a sample-efficient sub-sampling of tensor slices. Our sample efficient policy is learned via an adaptable stochastic Thompson sampling using Dirichlet distributions with conjugate priors.
\end{abstract}
\section{INTRODUCTION}
Data tensors of orders 2 and greater are routinely being generated. These data collections are increasingly huge and growing. For instance, a North American Regional Reanalysis (NARR) has been collecting 70 climate variables every 3 hours from 1979 to the present, and it is currently at a total size of 29.4 Terabytes \citep{mesinger2006north}. These data tensors are tensor fields in which each location of data contains important information. Directly accessing such large data tensor collections for information has become increasingly prohibitive. Approximate low rank and low memory representation (compact) generation of tensor sketches provide a space and time efficient alternative as all information querying and post-processing with high accuracy is performed on the sketches.

Randomized sketching algorithms have been a popular approach for producing a low rank approximation of very large matrices and tensors \citep{woolfe2008fast, halko2011finding,woodruff2014sketching,cohen2015dimensionality,boutsidis2016optimal,tropp2017practical}. The algorithms are faster than the singular value decomposition (SVD) while keeping the approximation accuracy similar to the SVD. Recently, SketchyCoreSVD has been proposed to further speed up the computation and reduce the memory requirement by building random sketches only from sub-sampled and projected columns and rows \citep{bajaj2019sketchycoresvd}. It advances SketchySVD \citep{tropp2019streaming} by exploiting redundancy in the data matrix, which can be characterized by incoherence. SketchySVD had been extended to compute the approximate low-rank Tucker decomposition of tensors \citep{sun2020low}. Using similar ideas as in SketchyCoreSVD \citep{bajaj2019sketchycoresvd},  we derive a  Randomized SketchyCoreTucker (R-SCT) decomposition of tensor fields and a randomized sketchy version of the Tensor Train decomposition (R-SCTT) in the Appendix.

R-SCT is mathematically proven and computationally advantageous, but its performance has an inherent variance due to the random sampling protocol and user-defined sub-sampling ratios. R-SCT randomly samples data subsets from the original data tensor, and thus it can be suboptimal and inefficient for sparse structured data. In many modern scientific data, meaningful information is often concentrated in small regions --- e.g., specific patterns of tissue in medical scans and extreme climate events such as storms or droughts in satellite climate data. Sampling irrelevant or duplicate subsets yields unnecessary memory consumption without accuracy improvement. Moreover, the performance of the method can be impacted by additional requirements of optimal sampling ratios given available computational resources. 

While randomized sampling based Tucker decompositions have been studied in the past, most algorithms sample {\it columns} of matrix unfoldings \citep{ahmadi2021randomized}. Also, most randomized sampling based algorithms utilize a prior multi-variate Gaussian distribution \citep{drineas2007randomized}, since these are theoretically the fastest. However, they are suboptimal over the sampling sequence and the total required sample size is inefficient. Other algorithms utilize measures such as leverage scores \citep{saibaba2016hoid,perros2015sparse,cheng2016spals} to infer an informativeness distribution. These algorithms use more information, but they are problematic for higher order and very large sized data tensors. 

To overcome several of the above challenges, we present a new method called {\it Progressive Sketching} that progressively samples {\it row} vectors of matrix unfolding of input tensor (i.e., order $K-1$ tensor slice of order $K$ input tensor) from the actively updated informativeness distribution. By doing so, we actively learn an optimal sub-sampling sequence policy for factored tensor sketches. By learning a sample efficient and near- optimal sub-sampling policy, one can then easily provide the best low rank and low memory approximation of the tensor fields with the smallest sample use.

Progressive SketchyCoreTucker sketching (P-SCT) produces more accurate,  low rank approximations than R-SCT using the same amount of input data subsets. In other words, P-SCT yields a sample efficient factored tensor sketch. Our progressive learned factored sketching can be applied to any randomized sketching based machine learning algorithm, and formulated as learned sequential decision making agents. In this paper we first use a Bayesian Thompson sampling framework. Our Thompson sampling framework can be extended  to a local uncertainty framework for contextual bandits \citep{Wang2020}. Deep reinforcement learning algorithms, such as Soft Actor-Critic \citep{Haarnoja2018,Haarnoja2018b,christodoulou2019soft} or Soft Q-learning with mutual information regularization \citep{grau-moya2018soft}, can be used for progressive sketching based on the algorithm introduced in this paper. We validate the performance of P-SCT on various large static and time-varying datasets while quantitatively comparing it to a class of randomized decomposition methods.

Overall, the main contributions  of this paper are the following: 
\begin{itemize}
    \item A new method of progressive sketching (P-SCT) applicable to tensor fields, to further optimize the performance of randomized sketching, by additionally learning an optimal Thompson sampling policy for tensor sketching, using Dirichlet distribution;
    \item P-SCT is a learned generation  of compact (low memory) and effective latent space embeddings.  Factored tensor can be universally utilized for several very large data driven tasks;
    \item By complexity analysis and empirical comparisons, we show how progressive sketching can further optimize the randomized sketching to produce accurate low rank approximations. Moreover, P-SCT utilizes a much smaller number of data fiber samples than R-SCT and other full-scan tensor factorization algorithms.
\end{itemize}

\section{BACKGROUND}\label{sec:background}
\subsection{Tucker Decomposition and Higher Order SVD (HOSVD)}\label{sec:hosvd}

Tucker decomposition has been considered as a multilinear generalization of SVD, and HOSVD is a constrained Tucker decomposition that ensures the orthogonality of factor matrices and all-orthogonality of the core tensor \citep{de2000multilinear}. For order 3 tensor $\mathcal{A}\in\mathbb{R}^{n_1\times n_2\times n_3}$, the {\it matrix unfolding} $\bm{A}_{(1)} \in \mathbb{R}^{n_1\times n_2n_3}$ contains the element $a_{i_1,i_2,i_3}$ at the position with row number $i_1$ and column number equal to $i_2 \times n_3 + n_2$. Other two matrix unfoldings $\bm{A}_{(2)}$ and $\bm{A}_{(3)}$ can be defined in the same manner. Then, we define {\it $k$-mode vectors}, defined as the $n_k$-dimensional vectors obtained from $\mathcal{A}$ by varying the index $i_k$ and keeping the other indices fixed. With this definition, the 1-mode product of a tensor $\mathcal{A}\in\mathbb{R}^{n_1\times n_2\times n_3}$ by a matrix $\bm{U}\in\mathbb{R}^{m_1\times n_1}$, denoted by $\mathcal{A}\times_1 \bm{U}$ is an $(m_1\times n_2 \times n_3)$-tensor of which the entries are given by $\sum_{i_1} a_{i_1,i_2,i_3}u_{j_1,i_1}$. Other two mode product $\mathcal{A}\times_2 \bm{U}$ and $\mathcal{A}\times_2 \bm{U}$ can be also defined accordingly. Finally, order 3 SVD can be stated as following Theorem \ref{thm:hosvd}.

\begin{theorem}[see \citep{de2000multilinear}]\label{thm:hosvd}
    \quad Every order 3 tensor $\mathcal{A}$ can be written as the product 
    \[
        \mathcal{A} = \mathcal{S}\times_1 \bm{U}^{(1)}\times_2 \bm{U}^{(2)}\times_3 \bm{U}^{(3)},
    \]
    in which $\bm{U}^{(k)}$ is a unitary $(n_k\times n_k)$-matrix; $\mathcal{S}$ is a $(n_1\times n_2\times n_3)$-tensor of which subtensors $\mathcal{S}_{i_k=\alpha}$, obtained by fixing the kth index to $\alpha$, have the property of all-orthogonality and ordering for all possible k.
\end{theorem}

The algorithm for calculating core tensor $\mathcal{S}$ and scaling matrices $\bm{U}^{(k)}$ is given in the Algorithm \ref{alg:hosvd}.

\begin{algorithm}[h]    
    \caption{HOSVD}
    \label{alg:hosvd}
\begin{algorithmic}
    \State {\bfseries Input:} $\mathcal{A}\in\mathbb{R}^{n_1\times n_2\times n_3}$
    \State {\bfseries Output:} $\mathcal{S}, \bm{U}^{(1)}, \bm{U}^{(2)}, \bm{U}^{(3)}$
    \For{$k=1$ {\bfseries to} $3$}
    \State $\bm{U^{(k)}}\gets r_k$ leading left singular vectors of $\bm{A}_{(k)}$
    \EndFor
    \State Set $\mathcal{S} = \mathcal{A} \times_1 \bm{U}^{(1)T} \times_2 \bm{U}^{(2)T} \times_3 \bm{U}^{(3)T}$
\end{algorithmic}
\end{algorithm}

\subsection{Thompson Sampling (TS)}
Thompson sampling utilizes online decision making where actions are taken sequentially in a manner that must balance between exploiting what is known to maximizing immediate performance and exploring to accumulate new information that may improve future performance \citep{russo2018tutorial,agrawal2012analysis,chapelle2011empirical,thompson1933likelihood}. 

\subsection{Uniqueness of R-SCT and P-SCT}
Ahmadi-Asl et al. (2021) categorized randomized algorithms for the Tucker decomposition and HOSVD into four groups: {\bf Random projection}~\citep{che2019randomized,minster2020randomized,zhou2014decomposition,sun2021tensor,wolf2019low,batselier2018computing,che2021randomized}, {\bf Sampling}~\citep{caiafa2010generalizing,sun2009multivis,oseledets2008tucker,saibaba2016hoid,tsourakakis2010mach,song2019relative,perros2015sparse,traore2019singleshot}, {\bf Random least-squares}~\citep{malik2018low}, and {\bf Count sketch}~\citep{wang2015fast,gittens2020adaptive,malik2018low,malik2020fast}. Numerical experiments showed that sampling group outperform random projection group in computation time with approximately the same results \citep{ahmadi2021randomized}. The random projection based algorithms are not suitable for huge tensors stored outside-of-cores since they need to pass the original data tensor multiple times. Randomized pass-efficient algorithms have been introduced to overcome this issue by minimizing the number of passes \citep{halko2011finding,woodruff2014sketching,boutsidis2016optimal,upadhyay2018price}, but none utilize the sampling approach for the solution. Randomized SketchyCoreTucker (R-SCT) is a unique algorithm that combines random projection and sampling to utilize the advantage of both approaches and overcome limitations rooted in using them separately. 

Many sampling-based algorithms compute each factor matrix by sampling the "{\it columns}" of the corresponding unfolding matrix or equivalently sampling the fibers of the original data tensor. However, our P-SCT samples the "{\it rows}" of the unfolding matrix or equivalently sampling the tensor slices. By doing so, P-SCT assures all sampled slices are intersected with each other and progressively updates the approximation of the core from the intersection of all sampled slices.

One main differentiation of our Progressive SketchyCoreTucker (P-SCT) is its sample efficiency earned by active learning the informativeness distribution utilizing the idea of Thompson sampling. Many sampling-based algorithms use different sampling protocols to minimize the number of samples and computation time, but none use active learning for progressiveness. Sampling based on leverage scores are proposed in \citep{saibaba2016hoid,perros2015sparse,cheng2016spals}. These algorithms are not sample efficient due to prior scanning entire original data tensor to compute the leverage scores. CUR decomposition can be considered a sampling technique with a difference that the sampling procedure is performed heuristically instead of randomly \citep{caiafa2010generalizing,oseledets2011tensor,saibaba2016hoid}. It is known that CUR approximations are less accurate than the SVD, and the quality of decomposition quite depends on selection of fibers. The optimal selection has been considered an NP-hard problem, and thus heuristic algorithms have been used for computing suboptimal solutions. Our progressive sketching approach can be applied to CUR decomposition to overcome this issue too. Recently, Song et al. (2019) extended matrix CUR decomposition to tensor CURT decomposition and proposed an randomized algorithm to compute a low-rank CURT approximation to a tensor \citep{song2019relative}. But this method does not use active learning to achieve sample efficiency.

Random least-squares and count-sketch groups are a randomized algorithms for ALS-based computation of the best mulitilinear rank approximation such as Higher-Order Orthogonal Iteration (HOOI) \citep{de2000multilinear}. A randomized least-squares HOOI algorithm is proposed in \citep{malik2018low}, but numerical experiment in \citep{ahmadi2021randomized} showed that sampling group have better computation efficiency. Count-sketch technique has been implemented in several algorithms, mostly for CANDECOMP/PARAFAC (CP) decomposition \citep{wang2015fast,gittens2020adaptive,malik2020fast}.

Other prior work on approximating tensor decompositions include Oh et al. (2018) which proposed P-T{\footnotesize UCKER}, a scalable Tucker decomposition that resolves the computational and memory bottleneck of ALS by row-wise updating factor matrices with parallelization \citep{oh2018scalable}. FREDE is a graph embedding based on matrix sketching that applies Frequent Directions --- a deterministic matrix sketching in the row-updates model ---  on a matrix-factorization interpretation of a neural embedding, VERSE \citep{tsitsulin2021frede}. Others introduced a randomized CP decomposition algorithms \citep{wang2015fast,gittens2020adaptive,malik2020fast}. The CP decomposition factorizes a tensor into a sum of component rank-one tensors~\citep{kolda2009tensor}, and it is a special case of Tucker decomposition with hyper-diagonal core tensor. Another type of tensor decomposition is the Tensor-Train (TT) decomposition \citep{oseledets2011tensor} which requires iterative matrix SVD computation, and thus SketchyCoreSVD or any randomized matrix SVD algorithm can be applied. Tucker, CP and TT decompositions produces different factored forms, each suitable for different generative  applications.
\section{PROGRESSIVE SKETCHING}

\subsection{Progressive SketchyCoreTucker (P-SCT)} \label{pskt}
P-SCT is for any order $K$ tensor field, so let us assume order 3 tensor $\mathcal{A} \in \mathbb{R}^{N_1 \times N_2 \times N_3}$ for explanation purposes, which is huge in size, and its spectrum is decaying fast enough for low rank approximation. Its matricizations along the three dimensions are denoted by $\bm{A}_{(1)}\in \mathbb{R}^{N_1\times(N_2 N_3)}$, $\bm{A}_{(2)}\in \mathbb{R}^{N_2\times(N_3 N_1)}$, $\bm{A}_{(3)}\in \mathbb{R}^{N_3\times(N_1 N_2)}$. Let us denote row space of $\bm{A}_{(k)}$ as $Row_k = \left\{\bm{A}_{(k)}^{(1,:)},\ldots,\bm{A}_{(k)}^{(N_k,:)}\right\}$. Then, $Row_k$ where $k=1,2,3$ are bandit arms of TS. As an action, we choose a $Row_k$ and sample from it without replacement. Note that a row vector of a matrix unfolding is a slice matrix in original geometric dimension of order 3 tensor. Thus, P-SCT samples order $K-1$ slice tensors to sketch an order $K$ tensor field. 

The amount of information contained in the samples ($\equiv$ order $K-1$ slice tensors) is the reward of the action. An empirical variance of data in the samples would be the first option for estimating the amount of information. However, it only provides us with the overall variation without considering directional variation along with each dimension. In order words, the variance ignores the different variations in spatial, temporal, and spectral dimensions of the data tensors. For tensor decomposition and our progressive sketching, this different information in different dimensions should be effectively and efficiently measured from the sampled data to learn the sampling policy in TS.

For Tucker decomposition, the information in different dimensions can be measured as the accumulated variation in data fibers along with $K-1$ dimensions. For instance, let us assume there is a sample $\bm{A}_{(1)}^{(\delta_1,:)}$ which is equivalent to the slice matrix $\mathcal{A}^{(\delta_1,:,:)}$. Both rows and columns (i.e., data fibers) of $\mathcal{A}^{(\delta_1,:,:)}$ are used in Tucker decomposition. We define the Sum of Absolute Differences ($SAD$) to capture the accumulated variation in these data fibers. The computation of SAD has a similar time complexity with the variance where $SAD$ of $\bm{A}_{(1)}^{(\delta_1,:)} \equiv \mathcal{A}^{(\delta_1,:,:)}$ is

\begin{equation}\label{eq:SAD}
\begin{split}
    \frac{1}{N_2N_3} \left( \sum_{j=1}^{N_3}\sum_{i=1}^{N_2 - 1} \left|\mathcal{A}^{(\delta_1,i,j)} - \mathcal{A}^{(\delta_1,i+1,j)}\right| + \right.\\
    \left.\sum_{i=1}^{N_2}\sum_{j=1}^{N_3 - 1} \left|\mathcal{A}^{(\delta_1,i,j)} - \mathcal{A}^{(\delta_1,i,j+1)}\right| \right).
\end{split}
\end{equation}

The absolute differences of data in data fibers of $\mathcal{A}^{(\delta_1,:,:)}$ are accumulated and normalized by the size of slice matrix. The normalization is required because what we want to measure is the amount of information per usage of the memory for the sample efficient Tucker decomposition. $SAD$ of samples from other $Row_k$ can be calculated in the same manner.

Dirichlet distribution is used to balance exploitation and exploration of $Row_k$. We update the concentration parameters of Dirichlet distribution by the entropy of normalized $SAD$ of samples from $Row_k$. By doing so, P-SCT samples more from $Row_k$ of which information is uniformly distributed rather than concentrated in a narrow region. It effectively reduces the mutual information between samples.

At first, we assume concentration parameters of the Dirichlet distribution as unity to make the distribution equivalent to the uniform distribution $\bm{\alpha} = \{\alpha_1,\alpha_2,\alpha_3\} = \{1,1,1\}$. Then, sampling ratios $\bm{p} = \{p_1,p_2,p_3\}$ are drawn from the \textit{Dirich} $(\bm{\alpha})$. By multipling the batch-size per round $N_{batch}$ to $\bm{p}$, we get the number of samples $\bm{n} = \{n_1,n_2,n_3\}$ that we newly sample from each $Row_k$.

We utilize $SAD$ once again here for the sampling. Rather than randomly sample, we assign weights by $SAD$. As a result, P-SCT searches samples that contain the highest amount of information. Combined two search schemes --- 1) TS with Dirichlet distribution that balances $Row_k$ selection minimizing the mutual information between samples and 2) $SAD$-weighted sampling that maximizes the use of information of samples --- is the core of P-SCT algorithm that learns optimal subsampling policy for streaming tensor sketching.

At the first round, we know nothing about the input tensor. Thus, weights are set as unity that can be denoted by $\bm{W} = \{\bm{w}_1,\bm{w}_2,\bm{w}_3\},\bm{w}_k = \{1,...,1\}, n(\bm{w}_k) = N_k.$ where $\bm{w_k}$ is weights for $Row_k$ and $n(\bm{w}_k)$ denotes the length of weights. Let us define sampled indices $\bm{\Delta} = \{\bm{\Delta}_1,\bm{\Delta}_2,\bm{\Delta}_3\}$ and unsampled indices $\bm{\Omega} = \{\bm{\Omega}_1,\bm{\Omega}_2,\bm{\Omega}_3\}$ and an allowed total number of samples $N_{allow}$. Then, the main steps of P-SCT are the following. In the algorithm, $\lfloor \cdot \rfloor$ represents the floor function.

\begin{enumerate}
    \item {\it Sample from $Row_k$}
    \begin{itemize}
        \item Draw sampling ratios $\bm{p} = \{p_1,p_2,p_3\}$ from \textit{Dirich} $(\bm{\alpha})$.
        \item For $Row_k$, sample $n_k = \lfloor p_kN_{batch} \rfloor$ indices from unsampled indices $\bm{\Omega}_k$ following weights $\bm{w}_k$. Denote the indices of the sampled rows to be $\bm{\Delta}_k$. 
    \end{itemize}
    
    \item Compute $SAD$ and update TS parameters
    \begin{itemize}
        \item Compute $SAD$ of $\bm{A}_{(k)}^{(\bm{\Delta}_k,:)}$.  Denote $SAD$ of $\bm{A}_{(k)}^{(\delta_k,:)}, \delta_k\in\bm{\Delta}_k$ as $SAD(\delta_k)$.
        \item Compute $SAD$ entropy $\mathcal{H}_k = -\sum_{\delta_k\in\Delta_k} \frac{SAD(\delta_k)}{\sum_{\delta_k} SAD(\delta_k)}\log{\left(\frac{SAD(\delta_k)}{\sum_{\delta_k} SAD(\delta_k)}\right)}.$
        \item Update the concentration parameter $\alpha_k \leftarrow \mathcal{H}_k$.
        \item Update the unsampled indices $\bm{\Omega}_k \leftarrow \bm{\Omega}_k\backslash\bm{\Delta}_k$.
        \item Linearly interpolate $SAD$ of unsampled rows $SAD(\bm{\Omega}_k)$ based on the index differences using $SAD(\bm{\Delta}_k)$.
        \item Set $SAD(\bm{\Delta}_k) = 0$  to avoid sampling duplicates, and update $\bm{w}_k$ by normalized $SAD$.
        \item Repeat Steps 1 and 2 until $n(\bm{\Delta}_1) + n(\bm{\Delta}_2) + n(\bm{\Delta}_3) > N_{allow}$.
    \end{itemize}
    
    \item {\it Final SketchyCoreTucker}
    \begin{itemize}
        \item Compute rank-$\bm{r}$ approximation using SketchyCoreTucker (see Supplementary). 
        \item Instead of uniformly sampling rows of $\bm{A}_{(1)}$,$\bm{A}_{(2)}$, $\bm{A}_{(3)}$, use $\bm{\Delta}_1$, $\bm{\Delta}_2$, $\bm{\Delta}_3$.
        \item For the columns of $\bm{A}_{(1)}$,$\bm{A}_{(2)}$, $\bm{A}_{(3)}$, use corresponding fibers crossing the intersection $\mathcal{A}^{(\bm{\Delta}_1,\bm{\Delta}_2,\bm{\Delta}_3)}$.
        \item Dimension reduction parameters $\bm{k}$ and $\bm{s}$ can be decided as 
        \[
            \bm{k} = \left\lfloor \bm{r} + \frac{1}{3}(\bm{n} - \bm{r}) \right\rfloor, \quad 
            \bm{s} = \left\lfloor \bm{r} + \frac{2}{3}(\bm{n} - \bm{r}) \right\rfloor,
        \]
        where $\bm{n} = \{n(\bm{\Delta}_1),n(\bm{\Delta}_2),n(\bm{\Delta}_3)\}$.
    \end{itemize}
\end{enumerate}

\subsection{Complexity Analysis}
Table \ref{table:complexity} shows the complexity of R-SCT, P-SCT, and full-scan algorithms (RP-HOSVD, R-STHOSVD, R-PET, R-ST, and R-HOID \citep{ahmadi2021randomized}), an ALS-based algorithm (RP-HOOI Ibid.), and SkethcyCore version of Tensor Train decomposition (R-SCTT) for the order 3 tensor case. For a simple comparison, we assume that $N_1=N_2=N_3=N$, $r_1=r_2=r_3=r$, $m_1=m_2=m_3=m$, $s_1=s_2=s_3=s$, and $k_1=k_2=k_3=k$. Also, it is assumed that the same $b=N_{batch}/3$ rows are sampled from each $Row_k$ every iteration for P-SCT. Thus, a product of total number of iteration $I$ and $b$ is $m$. Also, note that $N>m>r>s>k$.

\begin{table*}[h]
\caption{The computational complexity for decomposing order 3 data tensor. The computational benefits of SketchyCoreTucker depend on how much $m$ is smaller than $N$. If the required $m$ for target accuracy is close to $N$, SketchyCoreTucker has similar computational complexity to other algorithms. The main advantage of P-SCT is reducing $m$ for target accuracy sustaining computational complexity of decomposition close to $N^2$.}
\label{table:complexity}
\vskip 0.15in
\centering
\begin{tabular}{c|ccccc}
\hline
Algorithm & Projection/Sampling & QR & Core & Total\\
\hline
RP-HOSVD & $O(N^3r)$ & $O(Nr^2)$ & $O(N^3r)$ & $O(N^3r)$\\
RP-STHOSVD & \multicolumn{2}{c}{$O(N^4)$} & $O(N^3r)$ & $O(N^4)$\\
R-PET & $O(N^3s)$ & $O(N^2k)$ & $O(Nsk+s^3k)$ & $O(N^3s)$\\
R-ST & $O(r)$ & & $O(N^3r)$ & $O(N^3r)$\\
R-HOID & $O(N^3k)$ & & $O(N^3r)$ & $O(N^3r)$\\
RP-HOOI & $O(IN^3r)$ & $O(INr^2)$ & $O(IN^3r)$ & $O(IN^3r)$\\
R-SCTT &  &  &  & $O(N^2mk)$\\ 
R-SCT & $O(Nmk+m^3s)$ & $O(Nk^2)$ & $O(msk+r^3k)$ & $O(Nmk+m^3s)$\\
P-SCT & $O(N^2m+m^3s)$ & $O(Nk^2)$ & $O(msk+r^3k)$ & $O(N^2m+m^3s)$\\
\hline
\end{tabular} 
\vskip -0.1in
\end{table*}

\subsection{Theoretical Guarantees}
Our proofs are based on the framework of SketchyCoreSVD, as outlined in \citep{bajaj2019sketchycoresvd}, with modifications to accommodate varying numbers of random subsamples from each dimension of the data tensor.

Suppose $\bm{A} \in \mathbb{R}^{M\times N}$ has an SVD of the following form
\begin{equation}
\bm{U\Sigma V}^* =  \begin{bmatrix}\bm{U}_1 & \bm{U}_2\end{bmatrix} \begin{bmatrix}\bm{\Sigma}_1 & \\ & \bm{\Sigma}_2 \end{bmatrix} \begin{bmatrix}\bm{V}_1 & \bm{V}_2\end{bmatrix}^*
\end{equation}

where $[[\bm{A}]]_r = \bm{U}_1\bm{\Sigma}_1\bm{V}_1^*$ is the best rank $r$ approximation of $\bm{A}$ and $(\mu,\nu)$-incoherent.

SketchyCoreSVD proved that the basis computed from a randomly selected subset of the columns captures the range of $\bm{A}$ with bounded error
\begin{equation}
\begin{split}
|| \bm{A} - \bm{Q}\bm{Q}^*\bm{A} ||_F \leq (C_1(p,k,r) + 1) \cdot ||\bm{\Sigma}_2||_F +\\
C_2(p,k,r) \cdot ||\bm{\Sigma}_2||_2
\end{split}\label{eq:proof}
\end{equation}
with probability at least $1 - \frac{4}{r^3} - \frac{4}{k^3}$, where 
\begin{equation}
\begin{split}
C_1(p,k,r) &= \sqrt{\frac{6e^2}{p}} \cdot \frac{k}{k-r+1} \cdot k^{\frac{3}{k-r+1}},\\
C_2(p,k,r) &= \sqrt{\frac{36e^2}{p}} \cdot \frac{\sqrt{k \log k}}{k-r+1} \cdot k^{\frac{3}{k-r+1}},
\end{split}
\end{equation}
$k \geq r+4$ is sketch size, $n \geq 8 \mu r \log r$ is number of columns, and $p=n/N$. The same proof can be applied to matrix unfoldings $\bm{A}_{(k)}$ of data tensor $\mathcal{A}$ to show that a randomly selected subset of the columns captures the range of matrix unfoldings. 

The bounded error of rank $r$ approximation $||\bm{A} - [[\hat{\bm{A}}]]_r||_F$ also has been proved conditioned on Equation~\ref{eq:proof} in SkethcyCoreSVD. The algorithm samples the same number of rows and columns to capture the range and co-range of data matrix $\bm{A}$, but the theoretical guarantee has shown independently for each dimension of $\bm{A}$. Therefore, our proposed P-SCT also has a bounded error for the multi-linear $\bm{r}$ approximation with the different number of subsamples from each dimension. 

Note that the error bound in Equation~\ref{eq:proof} is a function of $C_1(p,k,r)$, $C_2(p,k,r)$, $||\bm{\Sigma}_2||_F$, and $||\bm{\Sigma}_2||_2$. For different dimensions of data tensor, the target rank $r$ is given, and both $C_1(p,k,r)$ and $C_2(p,k,r)$ decreases as $p$ and $k$ increase, i.e., as we sample more. If $||\bm{\Sigma}_2||_F$ and $||\bm{\Sigma}_2||_2$ of a dimension is greater than others, larger $p$ and $k$ (i.e., more samples) are required to ensure the error bound is tight as needed. This observation resembles the idea of P-SCT, where we can get sample-efficient rank-$\bm{r}$ approximation by balancing sampling between dimensions based on $SAD$ entropy. Therefore, P-SCT can have a tighter error bound than R-SCT as we demonstrate with empirical results in the following section.

\section{EXPERIMENTS}\label{sec:experiment}
The proper multilinear ranks, $\bm{r}$, for each dataset are identified from the scree plots for the modes calculated by HOSVD (see Algorithm \ref{alg:hosvd}). There are no strong guarantees on the performance of HOSVD; however, it is widely believed to produce an approximation usually quite close to the best multilinear rank approximation. At rank $r$ for mode $k$, scree plot can be computed by $ scree_k(r) = \frac{1}{\|\bm{A}_{(k)}\|_F^2} \sum_{i=r+1} \sigma_i^2(\bm{A}_{(k)})$. Scree plots are shown in the Appendix.

For a rank $\bm{r} = (r_1,r_2,r_3)$ approximation $[\![\mathcal{A}]\!]_{\bm{r}}$ to $\mathcal{A}$, we measure its approximation error as
\begin{equation}
    err = \frac{\|[\![\mathcal{A}]\!]_{\bm{r}} - \mathcal{A}\|_F^2}{\|\mathcal{A}\|_F^2}.
\end{equation}

To the best of our knowledge, P-SCT is the first algorithm that introduces active learning in the streaming Tucker decomposition of data tensors. Thus, our performance comparison has been made two-fold. First, we compare the low-rank approximation accuracy with several full-scan algorithms with the same target rank (RP-HOSVD, RP-STHOSVD, R-PET, R-ST, R-HOID, and RP-HOOI \citep{ahmadi2021randomized}), an ALS-based algorithm (RP-HOOI Ibid.), and SketchyCore version of Tensor Train decomposition algorithm (R-SCTT). We represent the distribution and mean of errors and computation times evaluated over 100 trials. The computation time reported does not include computing matrix unfoldings and the final approximation $[\![\bm{A}]\!]_r$ since we can avoid unfolding using index-wise access to the data tensor and, in practice, normally use the method only for calculating decompositions.

Second, we compare the sample efficiency of our progressive sketch generation. To fairly compare the performance of P-SCT by using the same size of partial data tensor and undecided parameters, the sampling ratio of R-SCT is randomly selected in every trial. Sampling numbers of rows for R-SCT are randomly generated, and other parameters are decided in the same way described in P-SCT algorithm Step 3. We compare how low rank approximation error decreases as the number of samples increases. We represent this in the learning curves evaluated in 100 trials. To emphasize the sample efficiency of our method, the mean approximation accuracy of other algorithms that use all data have been shown together. We also provide visual comparisons on the few snapshots of data tensors in the Appendix. The experiments are executed from Python on a 64-bit Microsoft machine with an Intel i7-12700H CPU and 32 GB of RAM. Our implementation is available at \url{https://github.com/CVC-Lab/ProgressiveSketching}.

\subsection{Cardiac Magnetic Resonance Imaging}
\begin{figure}[htb]
\begin{center}
\includegraphics[width=0.99\columnwidth]{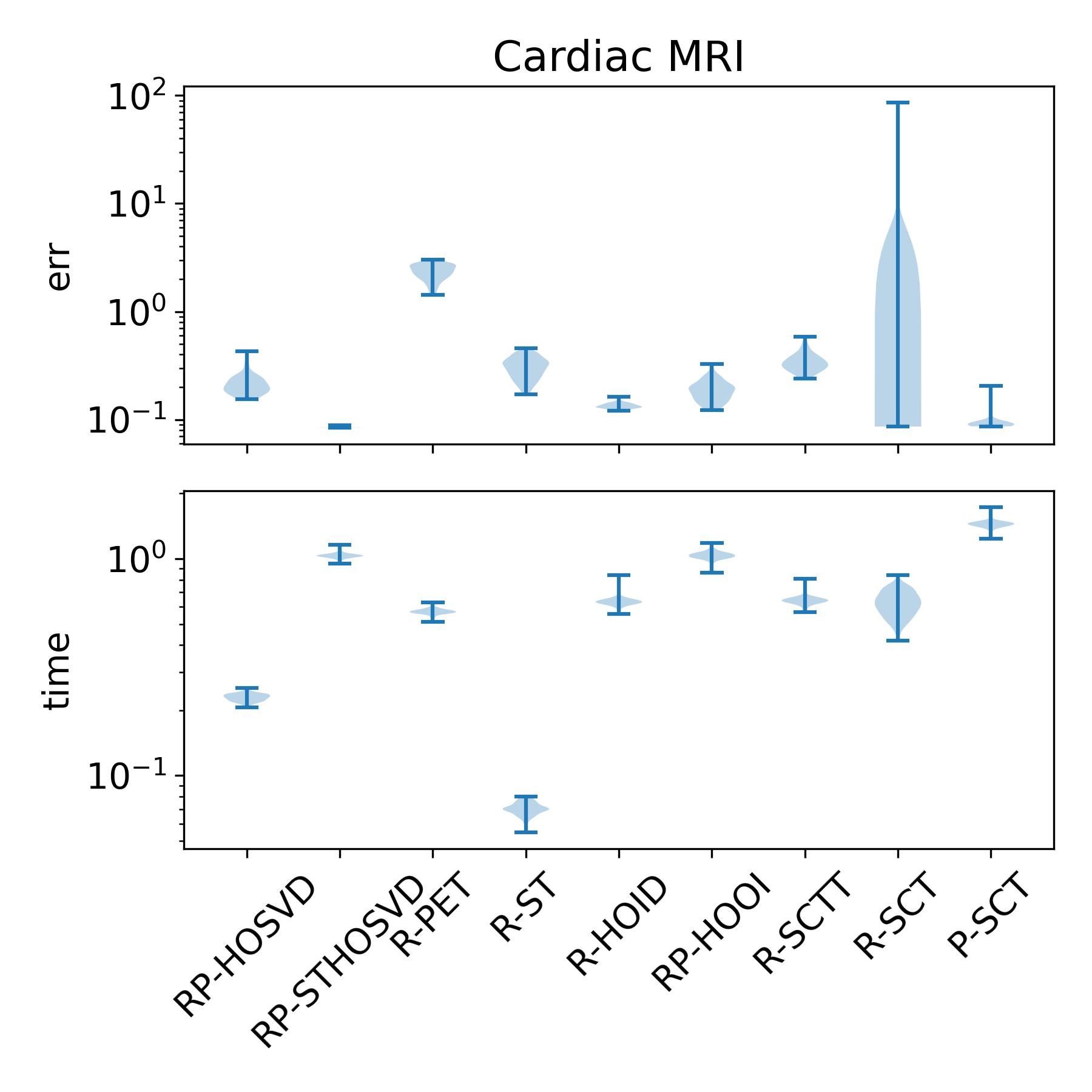}
\vskip -0.2in
\caption{Performance comparison for Cardiac MRI dataset. Violin plots of error and computation time (sec) from 100 trials. (see Table \ref{table:result} for more information).}
\label{fig:cardiac_result}
\end{center}
\end{figure}

\begin{figure}[htb]
\begin{center}
\includegraphics[width=\columnwidth]{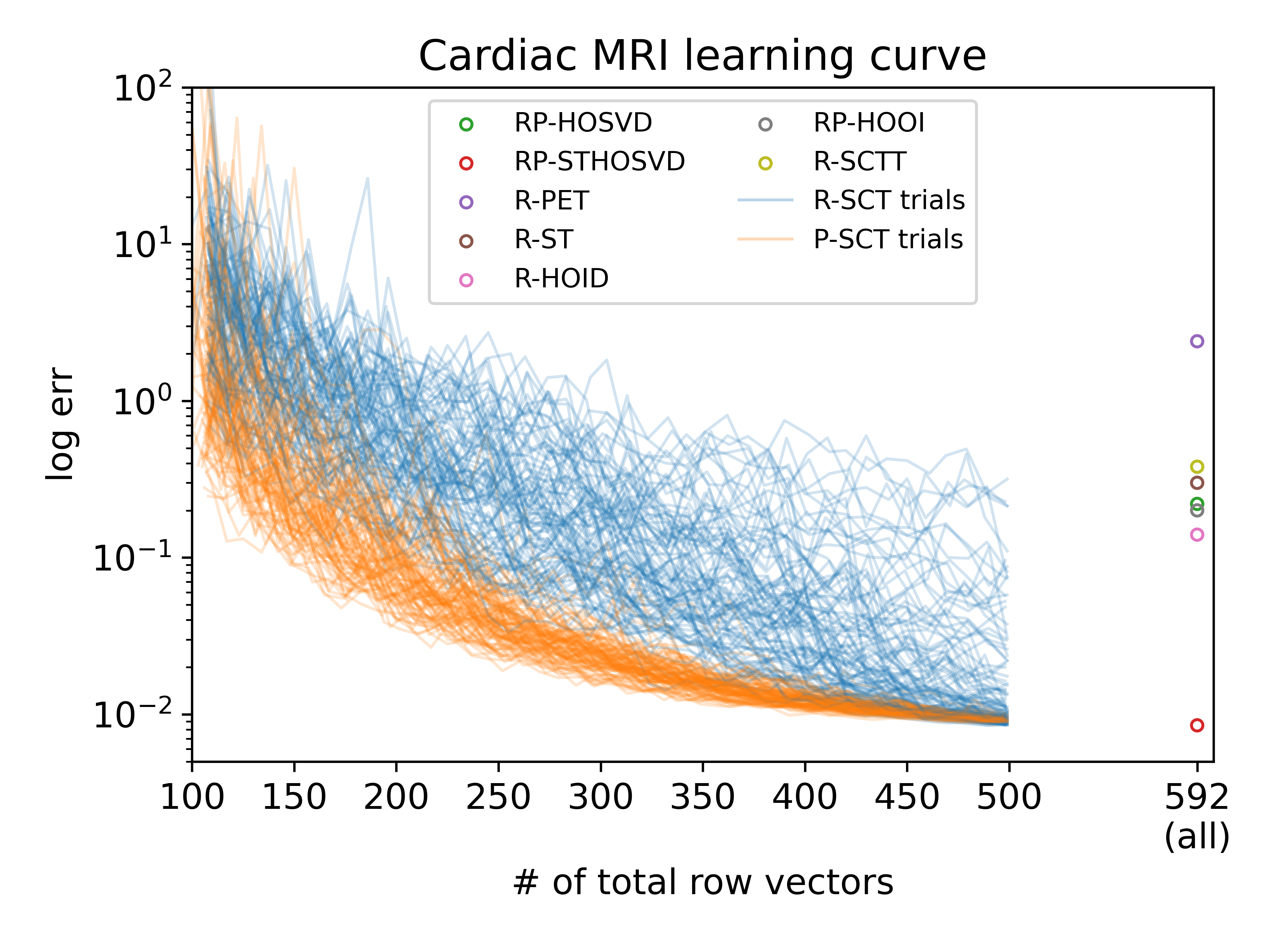}
\vskip -0.2in
\caption{Learning curve comparison. P-SCT (orange curves) converges faster than R-SCT (blue curves), showing sample-efficient behavior. P-SCT's error is smaller than R-SCT with smaller variances in almost every trial, empirically showing P-SCT's tighter error bounds. P-SCT produces a more accurate tensor decomposition than most full-scan algorithms using less than half of the input tensor.}
\label{fig:cardiac_lc}
\end{center}
\end{figure}

A collection of time-varying, 2D slice stacks of Cardiac MRI is tested: $\mathcal{A} \in \mathbb{R}^{256\times176\times160}$ consists of 160 time snapshots 2D images, each of size 256 $\times$ 176. The data is sparsely structured, where only a portion of the 2D slice contains the beating motion of the heart. Based on the scree plots shown in the Appendix, we choose $\bm{r} = (20,20,5)$. We also choose $N_{allow}=300$. Table \ref{table:result} and Figure \ref{fig:cardiac_result} show that P-SCT produces better performance than most full-scan algorithms only using approximately half of the data tensor. Figure \ref{fig:cardiac_lc} shows that P-SCT's performance converges faster than R-SCT to outperform most full-scan algorithms.

\begin{table*}[htb]
\caption{Performance comparison. Displayed values are mean $\pm$ standard deviation over 100 trials. Full-scan algorithms (RP-HOSVD, RP-STHOSVD, R-PET, R-ST, R-HOID) use all entries of the data tensor, but P-SCT uses only about half of the data tensor. Nevertheless, P-SCT produces good approximation results with acceptable computation time (sec). An ALS-based algorithm, RP-HOOI, also uses the original data tensor. Its accuracy can be adjusted by setting smaller error tolerance, but it substantially increases iteration and computational complexity. A SketchyCore version of tensor train decomposition, R-SCTT, randomly samples 50\% of input. RP-STHOSVD shows a better averaged performance, but the method is easily prohibitive as the size of the data tensor grows since its complexity is $O(N^4)$. Whereas, P-SCT is more scalable with a small sample size since its complexity is $O(N^2m+m^3s)$ where $N>m>s$.}
\label{table:result}
\vskip 0.15in
\centering
\begin{tabular}{|p{0.92in}|p{0.8in}|p{0.75in}|p{0.8in}|p{0.7in}|p{0.8in}|p{0.76in}|}
\hline
\multirow{2}{*}{ } & \multicolumn{2}{|c|}{Cardiac MRI} & \multicolumn{2}{|c|}{NARR Air Temperature}  & \multicolumn{2}{|c|}{Hyperspectral Image}  \\
\cline{2-7}
& err & time & err & time & err & time\\
\hline
RP-HOSVD   & 0.22 $\pm$ 5e-2  & 0.23 $\pm$ 9e-3  & 3.2e-5 $\pm$ 5e-6 & 1.3 $\pm$ 3e-2  & 8.7e-3 $\pm$ 2e-3 & 0.35 $\pm$ 4e-2\\
RP-STHOSVD & 0.085 $\pm$ 4e-4 & 1.0 $\pm$ 3e-2   & 8.2e-5 $\pm$ 2e-5 & 3.5 $\pm$ 0.2   & 3.2e-3 $\pm$ 3e-5 & 1.0 $\pm$ 9e-2\\
R-PET      & 2.4 $\pm$ 4e-1   & 0.57 $\pm$ 2e-2  & 3.9 $\pm$ 2e-1    & 2.7 $\pm$ 0.1   & 3.4 $\pm$ 5e-1    & 0.78 $\pm$ 8e-2\\
R-ST       & 0.32 $\pm$ 7e-2  & 0.071 $\pm$ 5e-3 & 5.6e-4 $\pm$ 8e-4 & 0.22 $\pm$ 2e-2 & 8.2e-3 $\pm$ 2e-3 & 6.0e-2 $\pm$ 1e-2\\
R-HOID     & 0.13 $\pm$ 8e-3  & 0.63 $\pm$ 3e-2  & 2.1e-5 $\pm$ 2e-6 & 2.9 $\pm$ 8e-2  & 7.4e-3 $\pm$ 4e-4 & 0.60 $\pm$ 7e-2\\
RP-HOOI    & 0.20 $\pm$ 4e-2  & 1.0 $\pm$ 4e-2   & 2.4e-5 $\pm$ 5e-6 & 2.6 $\pm$ 0.2   & 5.4e-3 $\pm$ 1e-3 & 0.57 $\pm$ 7e-2\\
R-SCTT     & 0.35 $\pm$ 7e-2  & 0.64 $\pm$ 3e-2  & 0.21 $\pm$ 2e-2   & 7.2 $\pm$ 0.1   & 0.14 $\pm$ 3e-2   & 1.9 $\pm$ 0.2\\
R-SCT      & 1.7 $\pm$ 9      & 0.63 $\pm$ 8e-2  & 0.10 $\pm$ 3e-1   & 1.6 $\pm$ 0.2   & 0.13 $\pm$ 0.2    & 0.59 $\pm$ 6e-2\\
P-SCT      & 0.094 $\pm$ 1e-2 & 1.4 $\pm$ 5e-2   & 2.4e-5 $\pm$ 5e-6 & 2.7 $\pm$ 0.1   & 1.3e-2 $\pm$ 2e-3 & 0.98 $\pm$ 0.1\\
\hline
\end{tabular}
\vskip -0.1in
\end{table*}

\subsection{NARR Air Temperature}
\begin{figure}[htb]
\begin{center}
\includegraphics[width=\columnwidth]{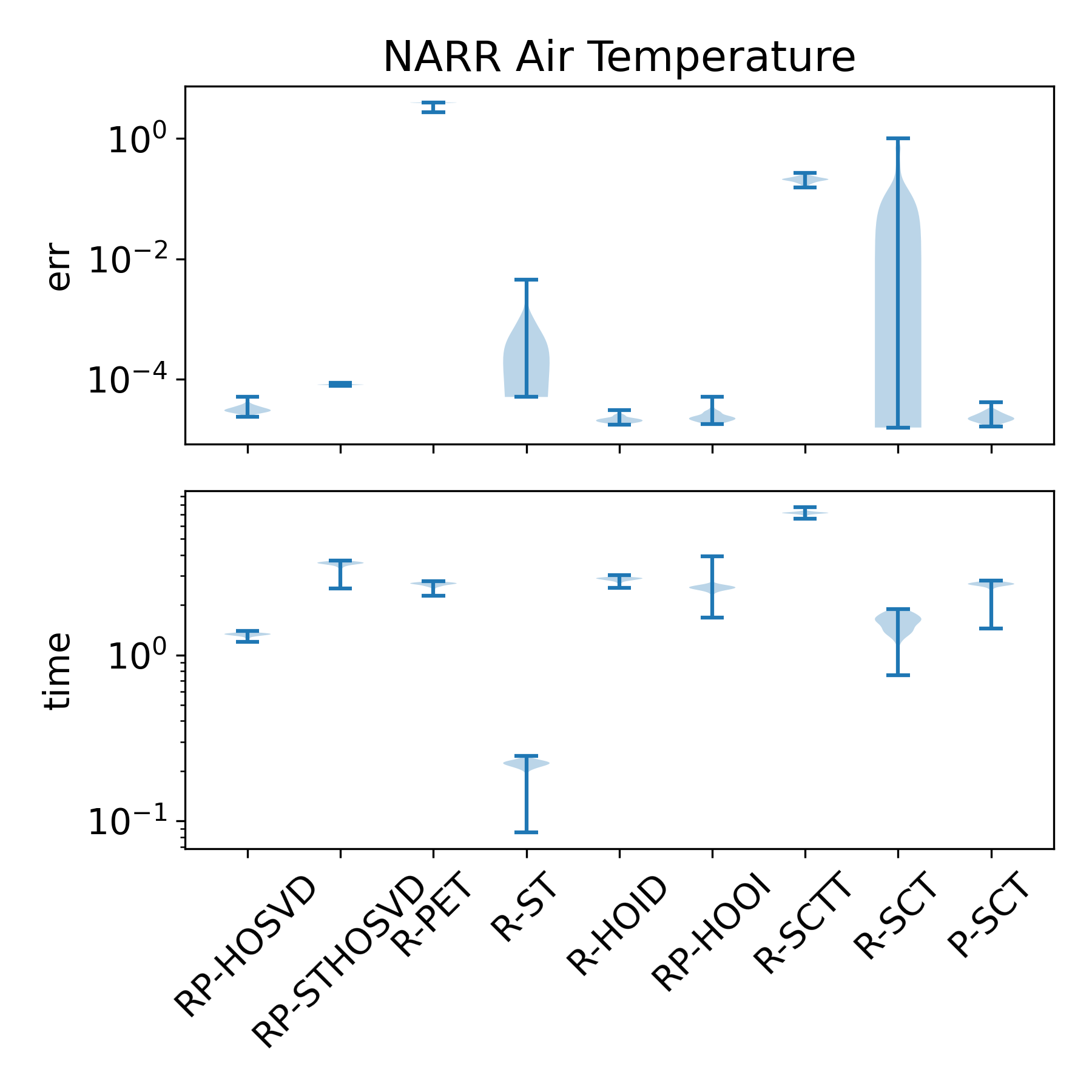}
\vskip -0.2in
\caption{Performance comparison for NARR air temperature dataset. Violin plots of error and computation time (sec) from 100 trials. (see Table \ref{table:result} for more information).}
\label{fig:air_result}
\end{center}
\end{figure}

\begin{figure}[htb]
\begin{center}
\includegraphics[width=\columnwidth]{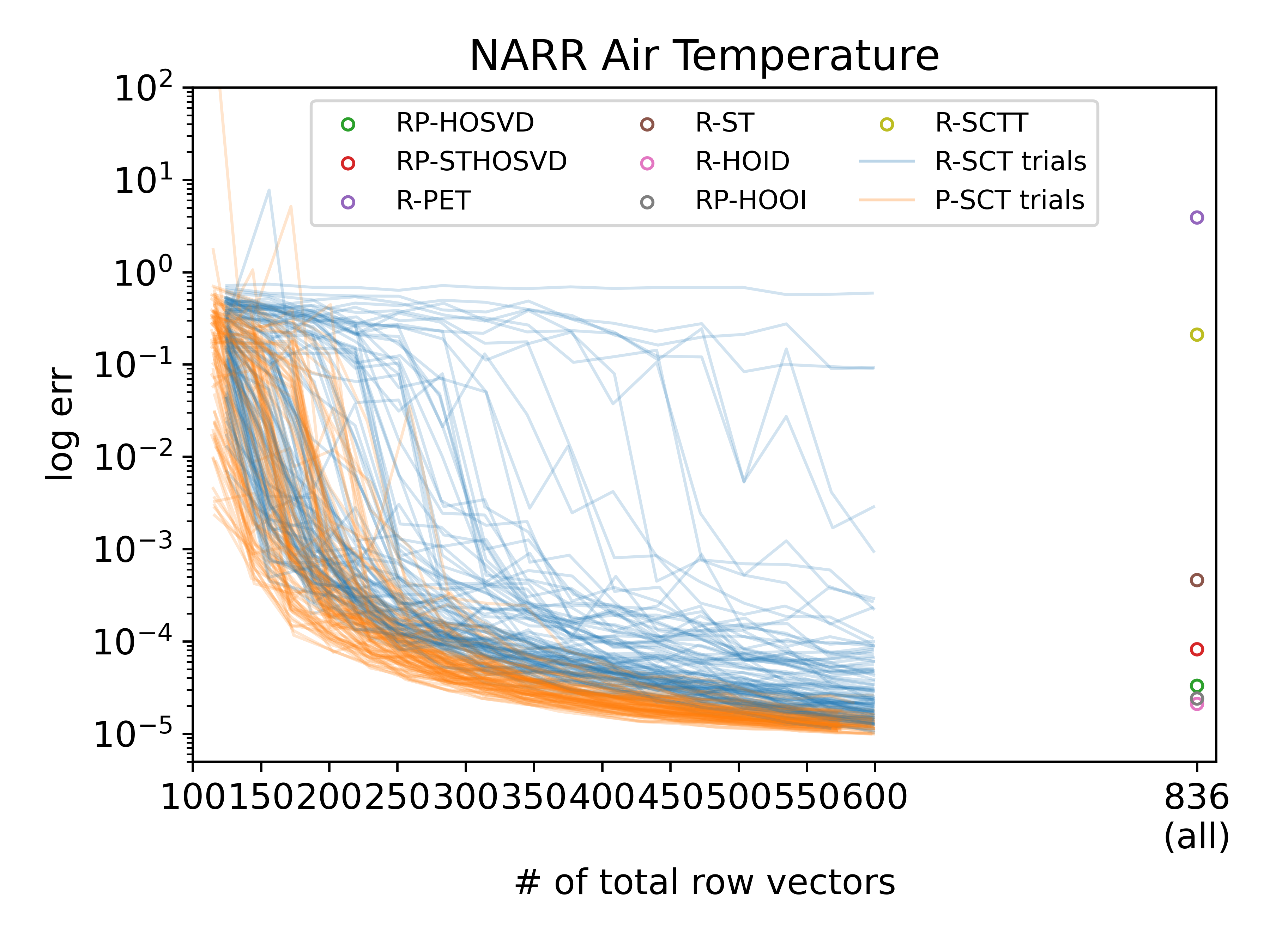}
\vskip -0.2in
\caption{Learning curve comparison. P-SCT (orange curves) converges faster than R-SCT (blue curves), showing sample-efficient behavior. P-SCT's error is smaller than R-SCT with smaller variances in almost every trial, empirically showing P-SCT's tighter error bounds. P-SCT produces a more accurate tensor decomposition than most full-scan algorithms using less than half of the input tensor.}
\label{fig:air_lc}
\end{center}
\end{figure}

NARR air temperature dataset is spatiotemporal series covering the North American continent \citep{mesinger2006north}. Its grid resolution is 349 $\times$ 277, which is approximately 0.3 degrees (32km) resolution at the lowest latitude. It provides 3 hourly series from 1979/01/01 to the present collected from 29 pressure levels (i.e., 29 different altitudes). We selected 1 month-long series in October 2020 at 500 hpa pressure level for the experiment. Spatial coverage is also clipped to discard null values around boundaries. As a result, $\mathcal{A}\in\mathbb{R}^{(248,252,336)}$. Based on scree plots shown in Appendix, we choose $\bm{r} = (50,10,20)$. $N_{allow}$ is assumed to be 400. The NARR air temperature is less sparsely structured than Cardiac MRI as wide regions' air temperature fluctuate differently. Still, Table \ref{table:result}, Figures \ref{fig:air_result} and \ref{fig:air_lc} show good results. P-SCT's performance converges faster to outperform some full-scan algorithms showing P-SCT's ability to find the important domain of the input tensor and selectively sample them for sample-efficient tensor sketching.

\subsection{Hyperspectral Image}
\begin{figure}[htb]
\begin{center}
\includegraphics[width=\columnwidth]{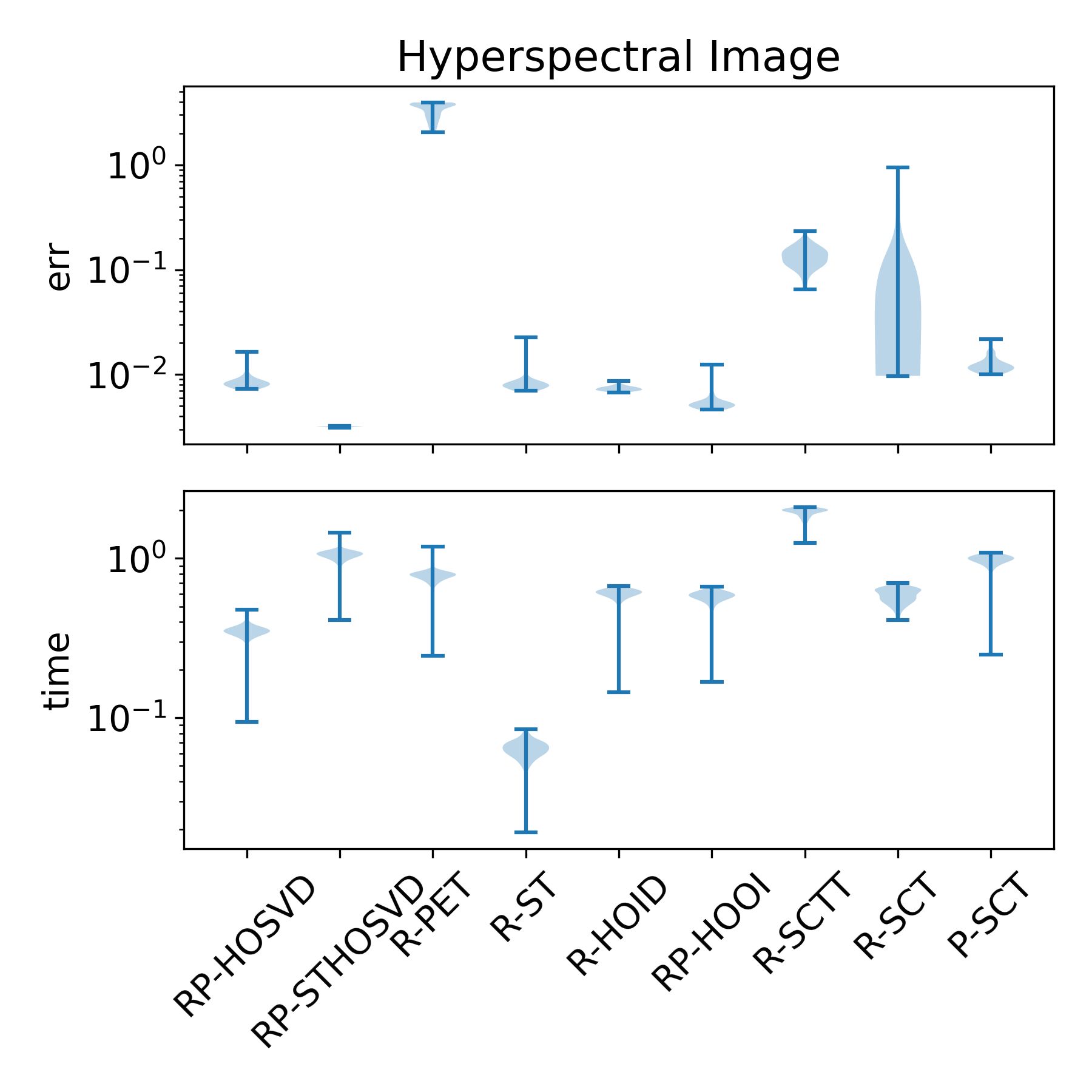}
\vskip -0.2in
\caption{Performance comparison for hyperspectral image. Violin plots of error and computation time (sec) from 100 trials. (see Table \ref{table:result} for more information).}
\label{fig:hsi_result}
\end{center}
\end{figure}

\begin{figure}[htb]
\begin{center}
\includegraphics[width=\columnwidth]{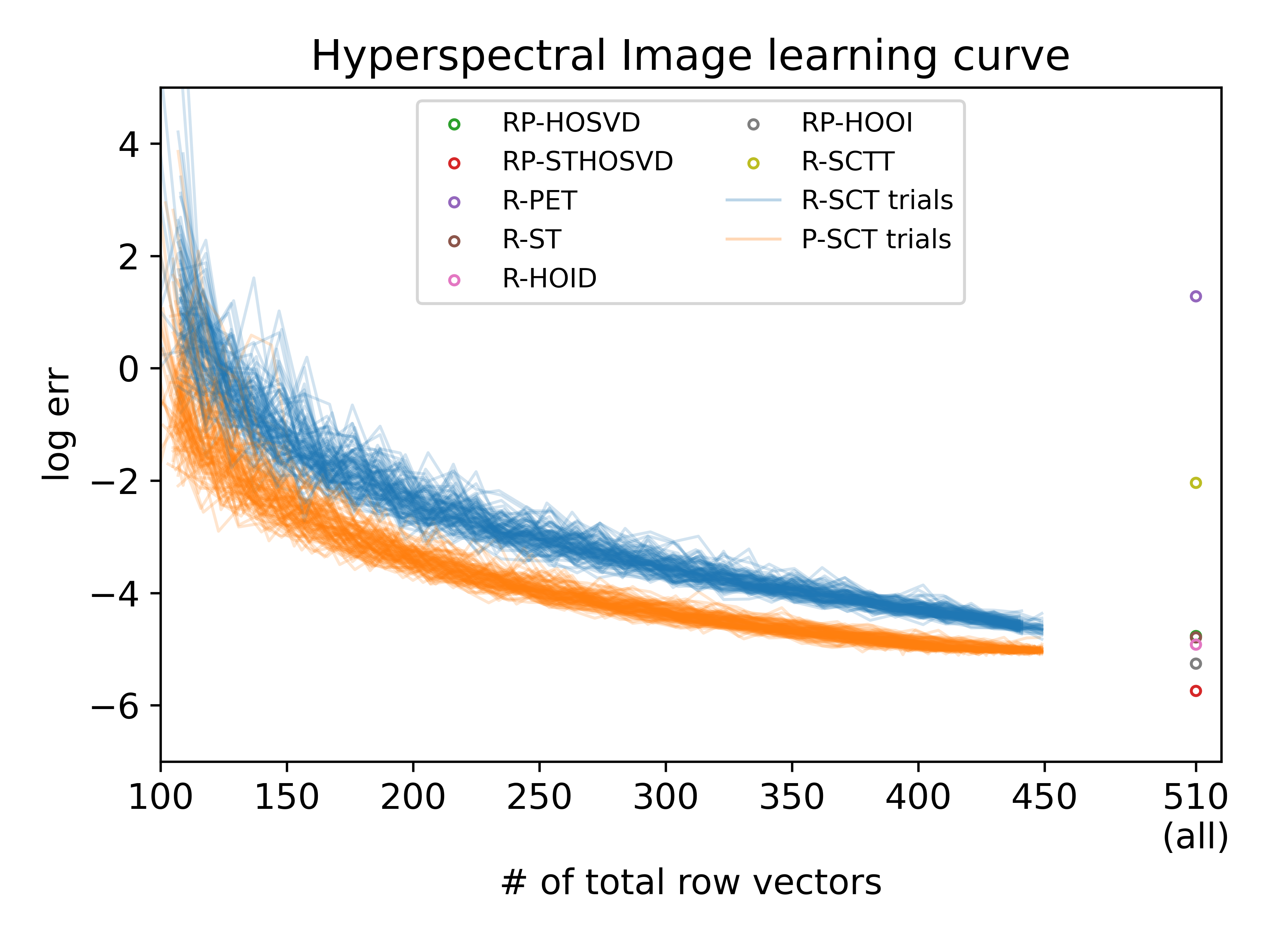}
\vskip -0.2in
\caption{Learning curve comparison. P-SCT (orange curves) converges faster than R-SCT (blue curves), showing sample-efficient behavior. P-SCT's error is smaller than R-SCT with smaller variances in almost every trial, empirically showing P-SCT's tighter error bounds. P-SCT produces a more accurate tensor decomposition than most full-scan algorithms using less than half of the input tensor.}
\label{fig:hsi_lc}
\end{center}
\end{figure}

Hyperspectral image (HSI) contains a wide spectrum of light instead of simple primary colors (e.g., red, green, blue) in each pixel with the purpose of finding objects or identifying materials. For example, the Airborne Visible / Infrared Imaging Spectrometer (AVIRIS) hyperspectral sensor data were acquired on June 12, 1992, over the Purdue University Agronomy farm northwest of West Lafayette and the surrounding area~\citep{PURR1947}. The data includes three approximately 2-mile by 2-mile test sites that were used to identify land use over major portions of the Indian Creek and Pine Creek watersheds. We selected one near the center (site 3) with the grid resolution 145 $\times$ 145 and 220 electromagnetic spectrum channels, and thus $\mathcal{A} \in \mathbb{R}^{145\times145\times220}$. Based on scree plots shown in the Appendix, we choose $\bm{r} = (25,25,5)$. $N_{allow}$ is assumed to be 300. HSI is sparsely structured, similar to NARR air temperature dataset, where the light spectrum is different for different land uses. Agian, Table \ref{table:result} and Figure \ref{fig:hsi_result} show that P-SCT produces a good performance as other full-scan algorithms, although it only uses about half of the data tensor. Figure \ref{fig:hsi_lc} compares progressive performance showing the sample efficiency of P-SCT.

\subsection{Discussion} \label{sec:discussion}
The computation time of P-SCT is longer than R-SCT. This delay is inevitable since the progressive sketching approach adds more computations. The delay will be greater for higher order tensor or bigger sized data tensor. However, considering a combinatorial number of possible sampling ratio parameters, the amount of computation time we trade off for getting optimal sample efficient low rank approximation is sufficiently negligible. The optimally balanced performance of the progressive sketching approach is well addressed that P-SCT maintains a good accuracy and computation time compared to fast-but-erroneous (RP-HOSVD, R-ST), and accurate-but-slow (RP-STHOSVD, R-HOID) full-scan algorithms.

\section{CONCLUSION}
\label{sec:conclusion}
Our paper presents various low-memory, latent factored tensor field sketches via a learned sub-sampling policy. P-SCT actively learns the optimal subsampling sketch streaming protocol so as to maximize the accuracy while minimizing the space requirements of the final tensor sketch.  Our numerical experiments show that P-SCT can get better performance with a limited number of samples than a random sampling based SketchyCoreTucker (R-SCT) and other full-scan algorithms (RP-HOSVD, RP-STHOSVD, R-PET, R-ST, R-HOID, and RP-HOOI from \citep{ahmadi2021randomized}). R-SCT and R-SCTT were derived from a generalization of a previously published SketchyCoreSVD method. While R-SCT and R-SCTT rely on a preset sampling ratio of the tensor fibers, P-SCT does not need this preset since the algorithm actively balances the ratio of tensor fibers from each dimension to get the maximum reward. P-SCT and in comparison to R-SCT and R-SCTT, also avoids further fine tuning that was required for SCT and TT. Further details on the progressive sketching version of TT (P-SCTT) and its comparison with R-SCTT are left for future work.

\subsubsection*{Acknowledgements}
This research was supported in part by a grant from the NIH DK129979, in part from the Peter O’Donnell Foundation, the Michael J. Fox Foundation, Jim Holland-Backcountry Foundation, and in part from a grant from the Army Research Office accomplished under Cooperative Agreement Number W911NF-19-2-0333. Additionally, we owe thanks to Tianming Wang for his help and insights when he was at UT Austin, and when this line of work was initiated. 

\bibliography{references}
\bibliographystyle{unsrtnat}

\clearpage

\pagebreak
\onecolumn
\aistatstitle{Supplementary Material for Sample Efficient Learning of Factored Embeddings of Tensor Fields}
\renewcommand\thesection{\Alph{section}}
\setcounter{section}{0}
\setcounter{algorithm}{0}
\setcounter{figure}{0}
\renewcommand{\thealgorithm}{\Alph{section}.\arabic{algorithm}} 
\renewcommand{\thefigure}{\Alph{section}.\arabic{figure}} 
\section{Randomized SketchyCore Tensor Decompositions}\label{sec:A}
\subsection{Randomized SketchyCoreTucker (R-SCT)}
SkechySVD has been extended to compute low-rank Tucker decomposition of tensors. One can call this method SketchyTucker. Using similar ideas as in SketchyCoreSVD, we derive a Randomized SketchyCoreTucker (R-SCT) method. 

Without loss of generality, we assume $\mathcal{A}\in\mathbb{R}^{N_1\times N_2\times N_3}$. Its matricizations along the three dimensions are denoted by $\bm{A}_{(1)}\in\mathbb{R}^{N_1\times(N_2N_3)}$, $\bm{A}_{(2)}\in\mathbb{R}^{N_2\times(N_3N_1)}$, and $\bm{A}_{(3)}\in\mathbb{R}^{N_3\times(N_1N_2)}$, respectively. 

Define $\bm{r}=[r_1,r_2,r_3],\bm{k}=[k_1,k_2,k_3],\bm{s}=[s_1,s_2,s_3],\bm{m}=[m_1,,m_2,m_3],\bm{n}=[n_1,n_2,n_3],$ where $\bm{r}\preceq\bm{k}\preceq\bm{s}\preceq\min\{\bm{m},\bm{n}\}$ and ``$\preceq$'' means ``$\leq$'' for each entry. Define $\mathbf{p}=[p_1,p_2,p_3]$ and $\mathbf{q}=[q_1,q_2,q_3]$ where $p_1=\frac{m_1}{N_2N_3}\in(0,1), p_2=\frac{m_2}{N_1N_3}\in(0,1), p_3=\frac{m_3}{N_1N_2}\in(0,1),$ and $ q_1=\frac{n_1}{N_1}\in(0,1), q_2=\frac{n_2}{N_2}\in(0,1), q_3=\frac{n_3}{N_3}\in(0,1).$

The main steps of R-SCT are summarized in Algorithm~\ref{alg:RSCT}.

\begin{algorithm}[htb]
	\caption{Randomized SketchyCoreTucker (R-SCT)}
	\textbf{Input:} $\mathcal{A}\in\mathbb{R}^{N_1\times N_2\times N_3}$, $\bm{r},\bm{k},\bm{s},\bm{p},\bm{q}$, uniformly sampled $m_i$ column indices of $\bm{A}_{(i)}$, $\Theta_i$, the map $\bm{\Omega}_i\in\mathbb{R}^{k_i\times m_i}$, uniformly sampled $n_i$ row indices of $\bm{A}_{(i)}$, $\Delta_i$, and the map $\bm{\Phi}_i\in\mathbb{R}^{s_i\times b_i}$ where $i=1,2,3$.\\
	\textbf{Output:} The near-optimal multi-linear $\bm{r}$ approximation, $$[[\mathcal{A}]]_{\bm{r}} = [[\mathcal{C}]]_{\bm{r}}\times_1\bm{Q}_1\times_2\bm{Q}_2\times_3\bm{Q}_3,$$ and the compact tensor sketch, i.e., a thumbnail of data tensor, $$[[\mathcal{C}]]_{\bm{r}}\times_1\bm{Q}_1^{(\Delta_1,:)}\times_2\bm{Q}_2^{(\Delta_2,:)}\times_3\bm{Q}_3^{(\Delta_3,:)}.$$
	\begin{algorithmic}[1]
        \State \textbf{Building randomized sketches}
		\State $\bm{Y}_1=\bm{A}_{(1)}^{(:,\Theta_1)}\bm{\Omega}_1^*\in\mathbb{R}^{N_1\times k_1};$ $\bm{Y}_2=\bm{A}_{(2)}^{(:,\Theta_2)}\bm{\Omega}_2^*\in\mathbb{R}^{N_2\times k_2};$ $\bm{Y}_3=\bm{A}_{(3)}^{(:,\Theta_3)}\bm{\Omega}_3^*\in\mathbb{R}^{N_3\times k_3};$
		\State $\mathcal{Z}=\mathcal{A}^{(\Delta_1,\Delta_2,\Delta_3)}\times_{1}\bm{\Phi}_1\times_{2}\bm{\Phi}_2\times_{3}\bm{\Phi}_3\in\mathbb{R}^{s_1\times s_2\times s_3};$
        \State \textbf{Compute the QR decomposition}
		\State $\bm{Y}_1=\bm{Q}_1\bm{R}_1,$ where $\bm{Q}_1\in\mathbb{R}^{N_1\times k_1}$, and $\bm{R}_1\in\mathbb{R}^{k_1\times k_1}$;
		\State $\bm{Y}_2=\bm{Q}_2\bm{R}_2,$ where $\bm{Q}_2\in\mathbb{R}^{N_2\times k_2}$, and $\bm{R}_2\in\mathbb{R}^{k_2\times k_2}$;
		\State $\bm{Y}_3=\bm{Q}_3\bm{R}_3,$ where $\bm{Q}_3\in\mathbb{R}^{N_3\times k_3}$, and $\bm{R}_3\in\mathbb{R}^{k_3\times k_3}$;
		\State $\mathcal{C}=\mathcal{Z}\times_1(\bm{\Phi}_1\bm{Q}_1^{(\Delta_1,:)})^{\dagger}\times_2(\bm{\Phi}_2\bm{Q}_2^{(\Delta_2,:)})^{\dagger}\times_3(\bm{\Phi}_3\bm{Q}_3^{(\Delta_3,:)})^{\dagger}\in\mathbb{R}^{k_1\times k_2\times k_3};$
		\State Compute the best multi-linear rank $\bm{r}$ approximation of $\mathcal{C}$, $[[\mathcal{C}]]_{\bm{r}}$, by an algorithm such as HOSVD or HOOI. 
	\end{algorithmic} \label{alg:RSCT}
\end{algorithm}

\clearpage

\subsection{Randomized SketchyCoreTensorTrain (R-SCTT)}
SketchyCoreTT computes the truncated SVDs in Step 3 and Step 7 in Algorithm~\ref{alg:TT} using SketchyCoreSVD. Besides $(k_1,s_1,k_2,s_2)$, there are two extra parameters  $(p_1,p_2)$ about sampling ratios.

\begin{algorithm}[htb]
	\caption{Tensor Train Decomposition}
	\textbf{Input:} $\mathcal{A} \in \mathbb{R}^{N_1 \times N_2\times N_3}$. \\
	\textbf{Output:} Tensor $\mathcal{B}$ in TT format with factor matrices $\bm{G}_1\in\mathbb{R}^{N_1\times r_1}$ and $\bm{G}_3\in\mathbb{R} ^{r_2\times N_3}$ and core $\mathcal{G}_2\in\mathbb{R}^{r_1\times N_2\times r_2}$.
	\begin{algorithmic}[1]
		\State Initialize $\mathcal{C}=\mathcal{A}$. 
		\State Set $\bm{C} = \text{reshape}(\mathcal{C},N_1,N_2N_3)$;
		\State Compute truncated SVD of $\bm{C}$ with rank $r_1$, denoted by $\bm{U}_1\bm{\Sigma}_1\bm{V}_1^*$;
		\State Set $\bm{G}_1=\text{reshape}(\bm{U}_1,N_1,r_1)$;
		\State Update $\bm{C} = \bm{\Sigma}_1\bm{V}_1^*$;
		\State Set $\bm{C} = \text{reshape}(\bm{C},r_1N_2,N_3)$;
		\State Compute truncated SVD of $\bm{C}$ with rank $r_2$, denoted by $\bm{U}_2\bm{\Sigma}_2\bm{V}_2^*$;
		\State Set $\mathcal{G}_2=\text{reshape}(\bm{U}_2,r_1,N_2,r_2)$;
		\State Set $\bm{G}_3 = \bm{\Sigma}_2\bm{V}_2^*$.
	\end{algorithmic} \label{alg:TT}
\end{algorithm}
\section{Supplementary figures}\label{sec:B}
\subsection{P-SCT Algorithm}
\begin{figure}[htb]
\begin{center}
\centerline{\includegraphics[width=0.9\textwidth]{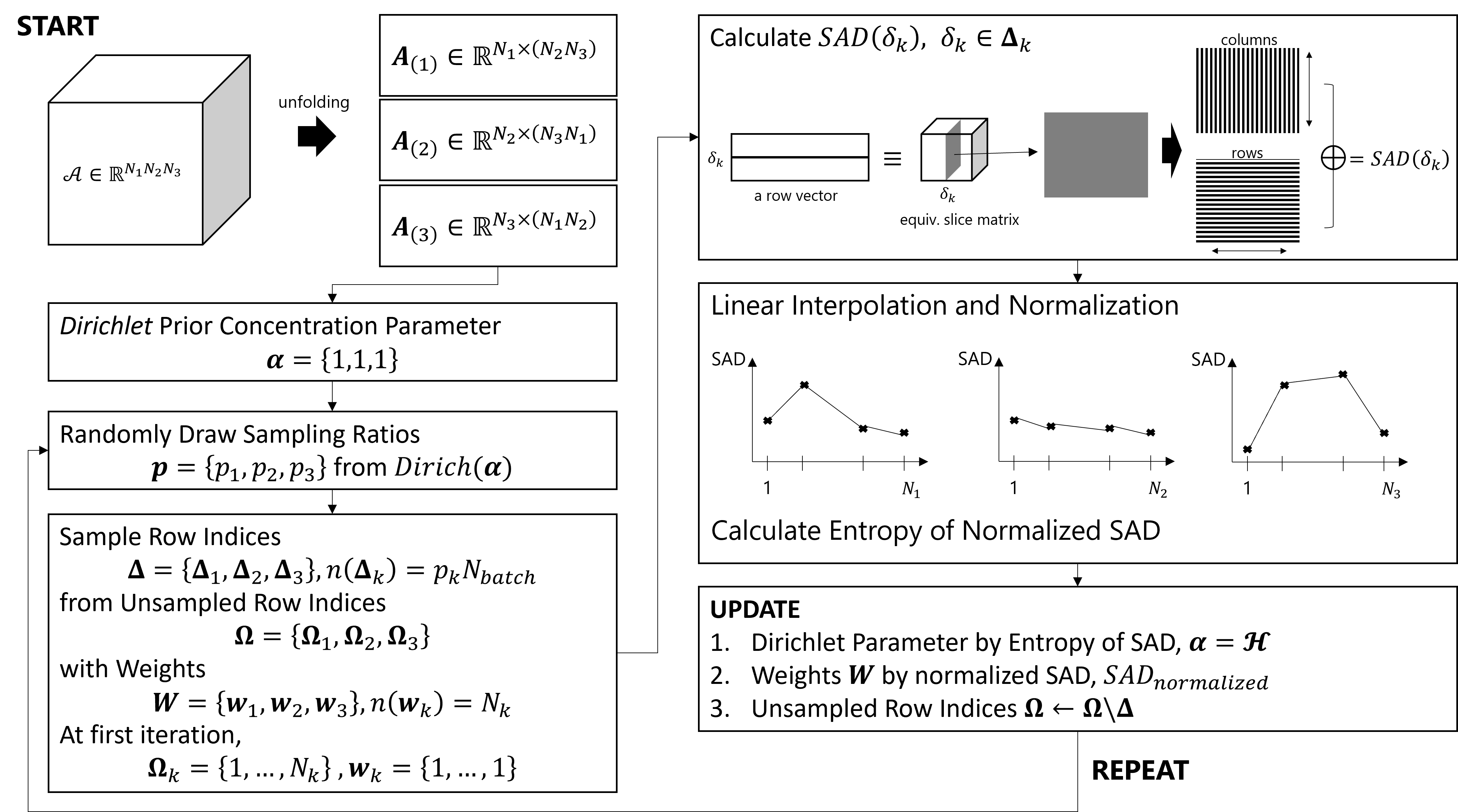}}
\caption{Flowchart of P-SCT algorithm. We draw sampling ratios from Dirichlet prior to balancing the exploration/exploitation of each row space of the matrix unfolding. The sum of Absolute Difference (SAD) is computed for each sampled tensor slice. The concentration parameter of Dirichlet prior and sampling weights are updated based on the SAD score and SAD entropy. By repeating this Thompson sampling scheme, P-SCT selectively samples informative regions in the input tensor.}
\label{fig:flowchart}
\end{center}
\end{figure}

Figure~\ref{fig:flowchart} illustrates the P-SCT algorithm.

\subsection{Scree plots}
The proper multilinear ranks, $\bm{r}$, for each dataset are identified from the scree plots for the modes calculated by HOSVD. At rank $r$ for mode $k$, scree plot can be computed by 
\begin{equation}
scree_k(r) = \frac{1}{\|\bm{A}_{(k)}\|_F^2} \sum_{i=r+1} \sigma_i^2(\bm{A}_{(k)}).
\end{equation}

\begin{figure}[H]
\begin{center}
\centerline{\includegraphics[width=0.85\textwidth]{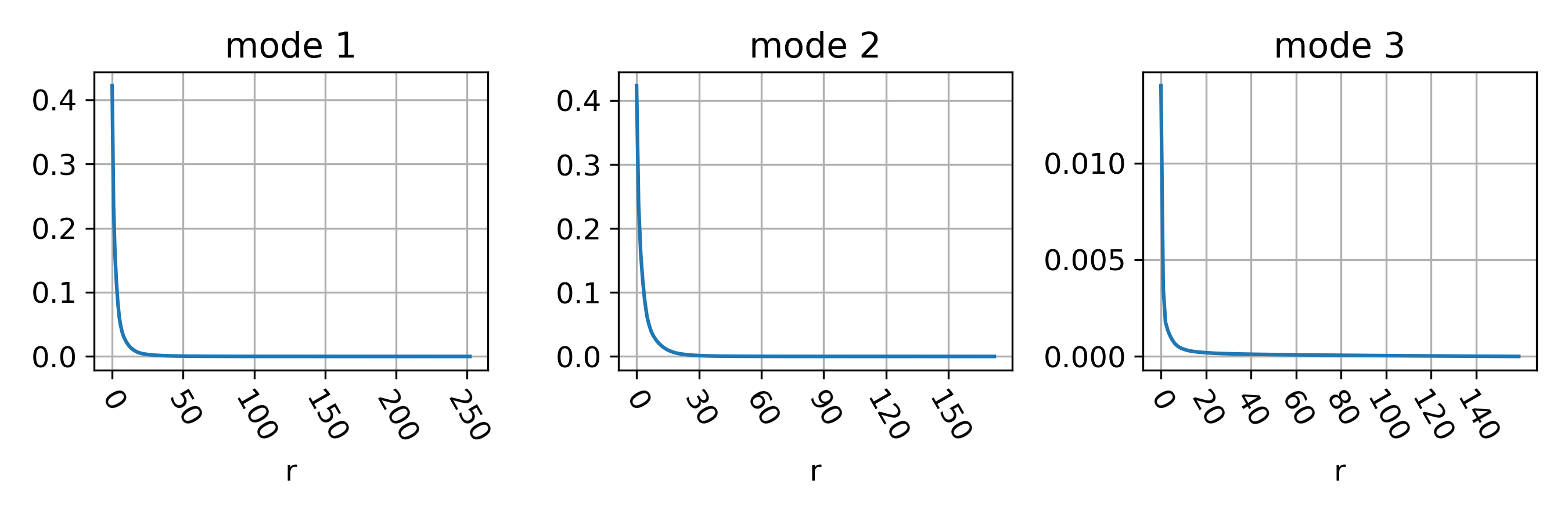}}
\vskip -0.2in
\caption{Scree plots for the modes of Cardiac MRI dataset. $\bm{r} = (20,20,5)$ has been selected for the low rank approximation.}
\end{center}
\end{figure}

\begin{figure}[H]
\begin{center}
\centerline{\includegraphics[width=0.85\textwidth]{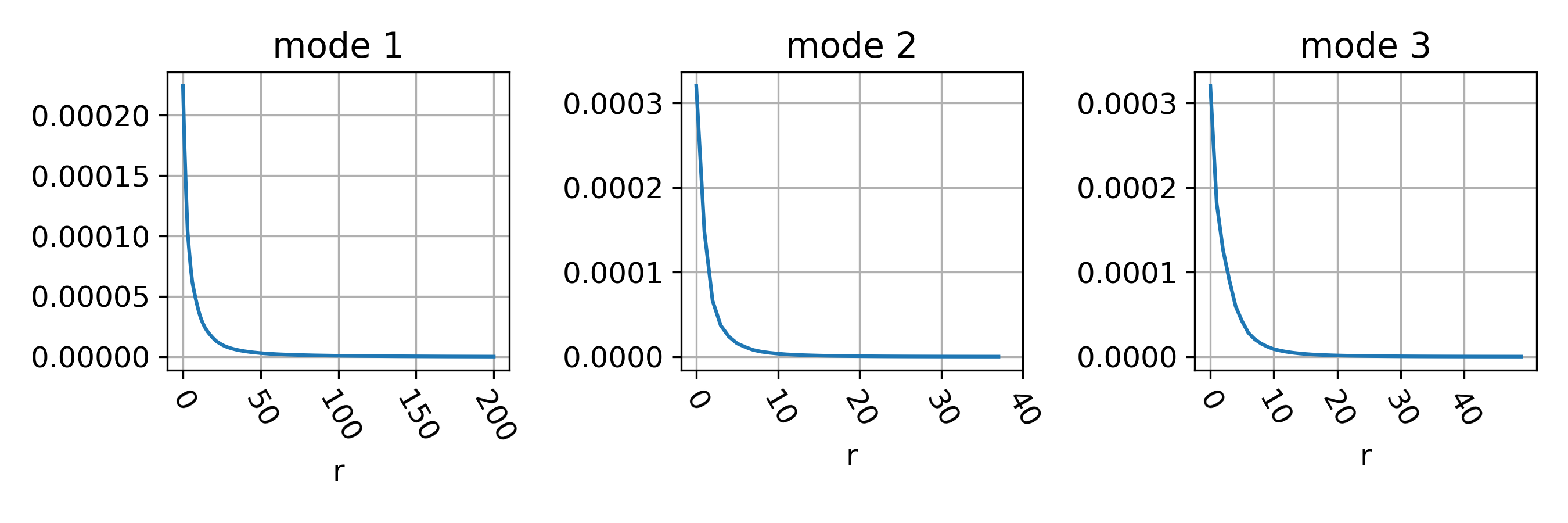}}
\vskip -0.2in
\caption{Scree plots for the modes of NARR air temperature dataset. $\bm{r} = (50,10,20)$ has been selected for the low rank approximation.}
\end{center}
\end{figure}

\begin{figure}[H]
\begin{center}
\centerline{\includegraphics[width=0.85\textwidth]{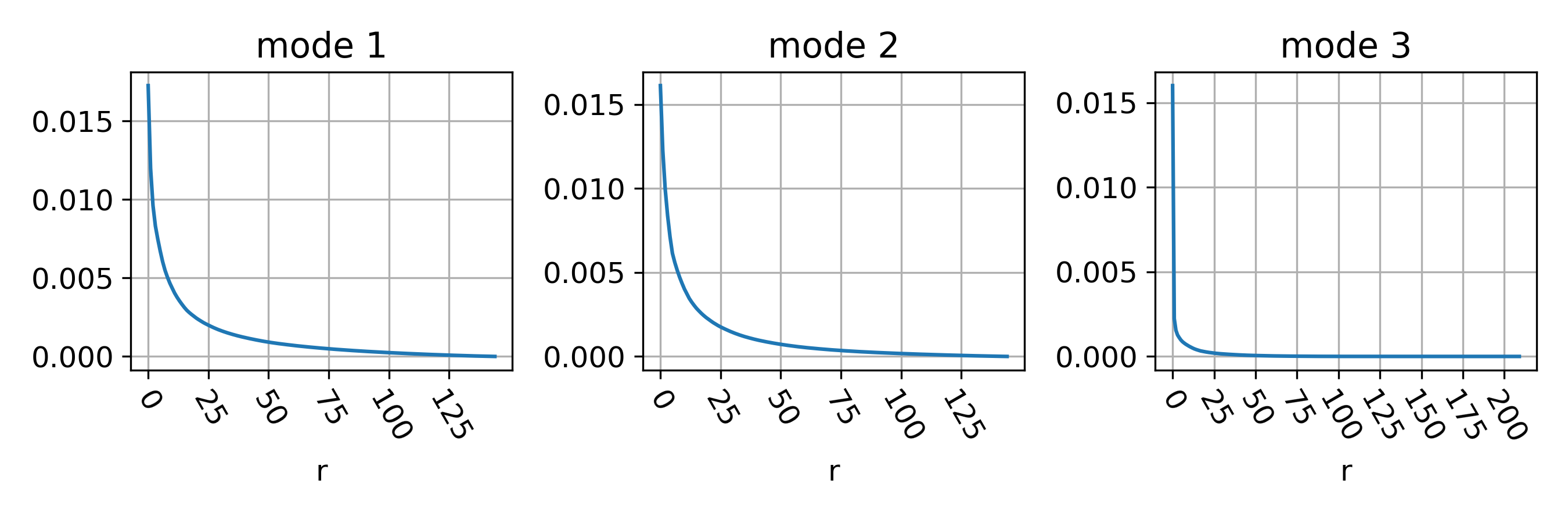}}
\vskip -0.2in
\caption{Scree plots for the modes of hyperspectral image. $\bm{r} = (25,25,5)$ has been selected for the low rank approximation.}
\end{center}
\end{figure}

\subsection{Snapshots}
\begin{figure}[htb]
\begin{center}
\centerline{\includegraphics[width=0.9\textwidth]{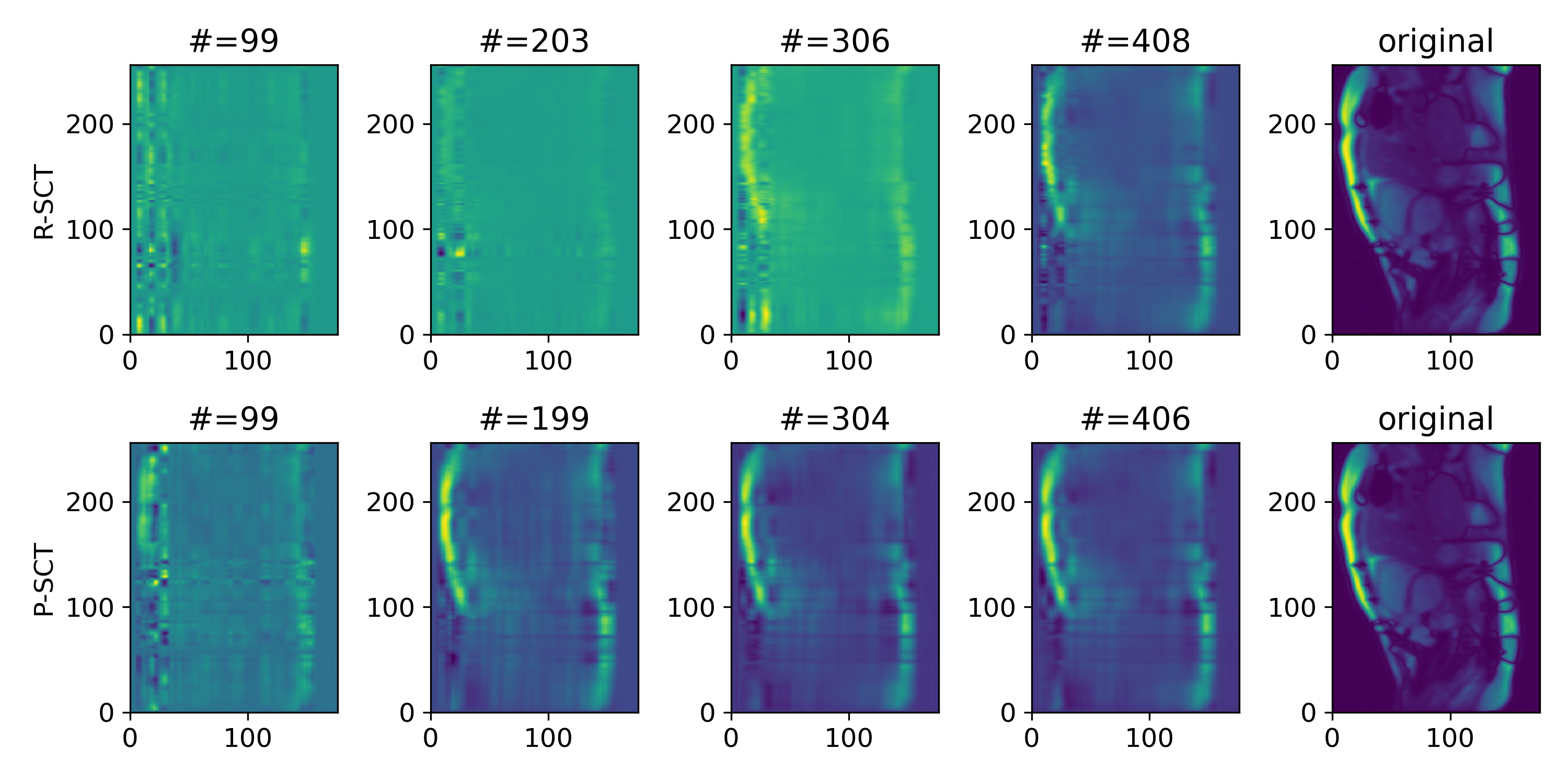}}
\caption{Cardiac MRI snapshots at the first timestamp.}
\label{fig:cardiac_Ar}
\end{center}
\end{figure}

\begin{figure}[htb]
\begin{center}
\centerline{\includegraphics[width=0.9\textwidth]{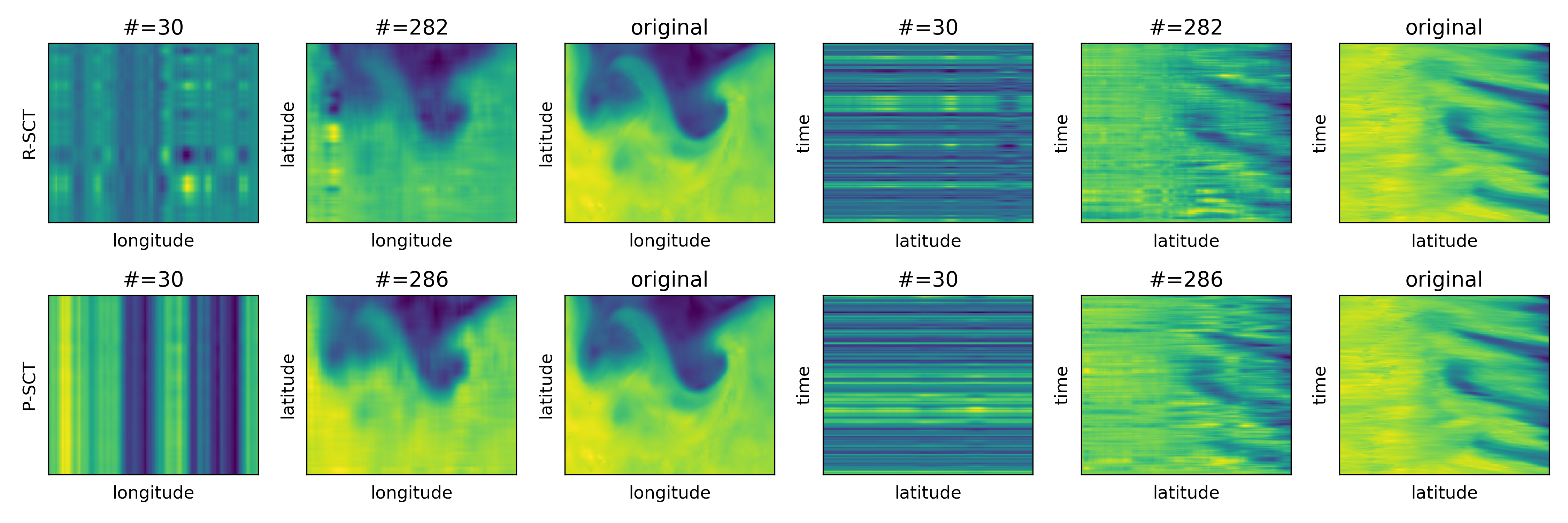}}
\caption{NARR air temperature snapshots at the first timestamp.}
\label{fig:air_Ar}
\end{center}
\end{figure}

\begin{figure}[htb]
\begin{center}
\centerline{\includegraphics[width=0.9\textwidth]{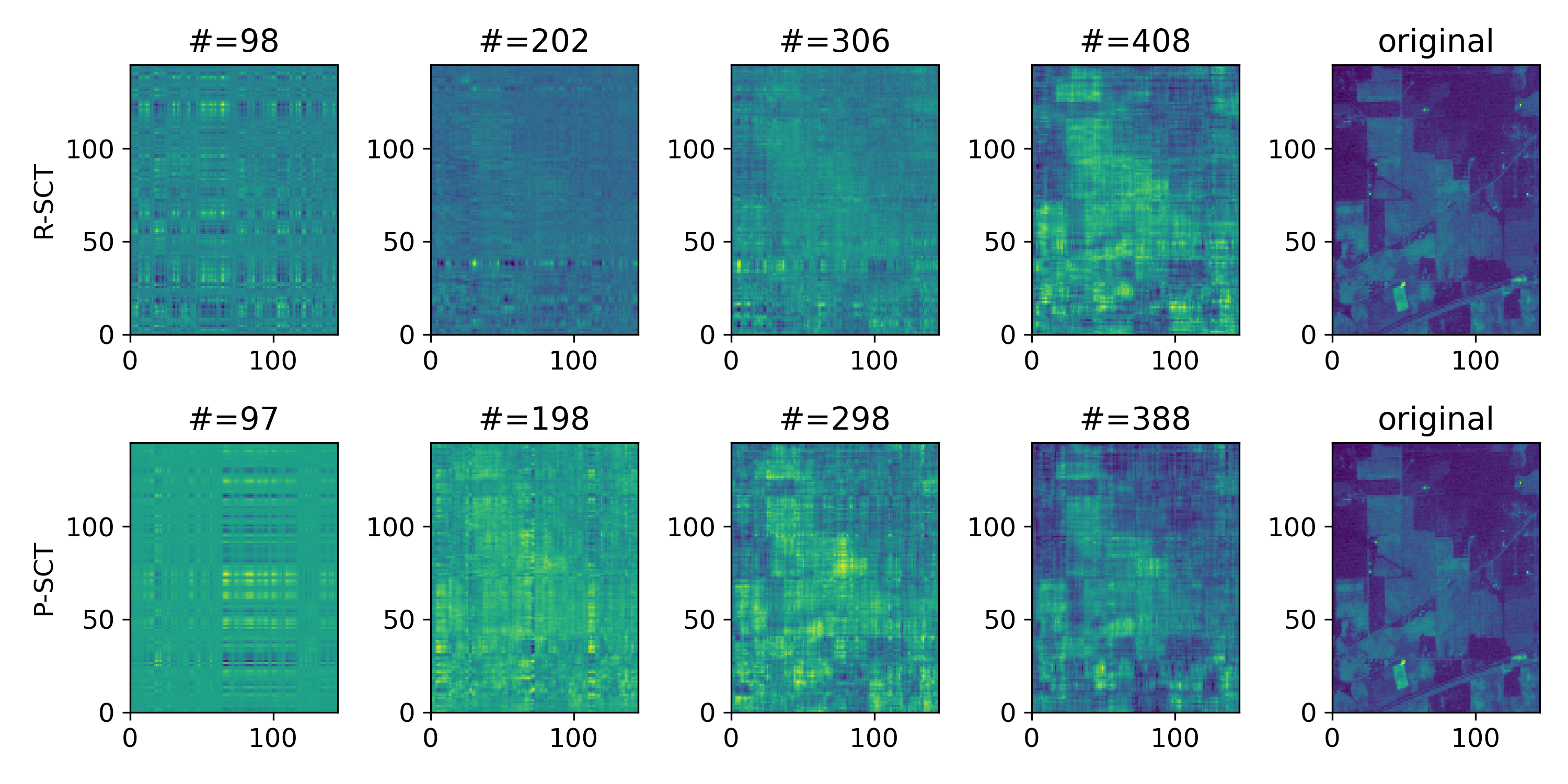}}
\caption{Hyperspectral image snapshots of the fourth band.}
\label{fig:hsi_Ar}
\end{center}
\end{figure}

Figures \ref{fig:cardiac_Ar}, \ref{fig:air_Ar}, and \ref{fig:hsi_Ar} show progressive visual comparison of the snapshot of reconstructed data tensors. Low rank approximation results from R-SCT and P-SCT using the number of samples indicated in each subtitle. For all cases, P-SCT reveals the data structure faster than R-SCT.

\end{document}